\documentclass[10pt,twocolumn,twoside]{IEEEtran}
\usepackage{amsmath}
\usepackage{amssymb}
\usepackage{float}
\usepackage{multirow}
\usepackage{graphicx}

\newtheorem{thm}{Theorem}

\newtheorem{defn}{Definition}

\newtheorem{lem}{Lemma}

\floatstyle{ruled}
\newfloat{algorithm}{tbp}{loa}
\providecommand{\algorithmname}{Algorithm}
\floatname{algorithm}{\protect\algorithmname}
\usepackage{algorithm}
\usepackage{algorithmic}

\begin{document}

\title{On Convergence Rate of the Gaussian Belief Propagation Algorithm for Markov Networks}
\author{{\normalsize{Zhaorong Zhang$^1$ and Minyue Fu$^{1}$}} 
\thanks{$^1$School of Electrical Engineering and Computer Science, The University of Newcastle. University Drive, Callaghan, 2308, NSW, Australia.}
\thanks{E-mails: zhaorong.zhang@uon.edu.au; minyue.fu@newcastle.edu.au.}
}
\maketitle
\begin{abstract}
Gaussian Belief Propagation (BP) algorithm is one of the most important distributed algorithms in signal processing and statistical learning involving Markov networks. It is well known that the algorithm correctly computes marginal density functions from a high dimensional joint density function over a Markov network in a finite number of iterations when the underlying Gaussian graph is acyclic. It is also known more recently that the algorithm produces correct marginal means asymptotically for cyclic Gaussian graphs under the condition of walk summability. This paper extends this convergence result further by showing that the convergence is exponential under the walk summability condition, and provides a simple bound for the convergence rate.  
\end{abstract}

\begin{IEEEkeywords} Gaussian belief propagation, belief propagation, Markov networks, Distributed algorithm, distributed estimation. 
\end{IEEEkeywords}

\section{Introduction}\label{sec1}
Belief Propagation (BP) algorithm is a well-celebrated distributed algorithm for Markov networks that has been widely utilised in many disciplines, ranging from statistical learning and artificial intelligence to distributed estimation, distributed optimisation, networked control and digital communications \cite{Pearl}-\cite{Tan}.  

Initially introduced by Pearl \cite{Pearl} in 1988, the BP algorithm is also known as {\em Pearl's algorithm}, {\em message-passing algorithm} and {\em sum-product algorithm}. It is designed to compute the marginal probability densities of random variables from the joint probability density function over a large Markov network with sparse connections among individual random variables. The significance of the algorithm stems from the facts that it is fully distributed (i.e., only local information is needed for iteration computation) and that a wide range of application problems can be formulated as a BP problem. It is well known that the BP algorithm produces correct marginal probability densities in a finite number of iterations when the underlying graph for the joint density function is acyclic (i.e., no cycles or loops). But the properties of the algorithm for cyclic (loopy) graphs have been a major research topic over several decades. 

The Gaussian BP algorithm, a special version of the BP algorithm for Markov networks with Gaussian distributions (also known as Gaussian graphical model), has received special attention for the study the convergence properties of Gaussian BP. In \cite{Weiss}, it was shown that Gaussian BP produces asymptotically the correct marginal means under the assumption that the joint information matrix is diagonal dominance. It was relaxed in \cite{Mailoutov} that the same asymptotic convergence holds when the joint information matrix is walk-summable, which is equivalent to the condition of generalised diagonal dominance. In \cite{Su,Su1},  necessary and sufficient conditions for asymptotic convergence of the Gaussian BP algorithm are studied. 

The purpose of this paper is to study the convergence rate of the Gaussian BP algorithm. Under the walk summability condition, we provide a simple bound for the exponential convergence rate of the marginal means. This bound is simply the spectral radius of the matrix related to the information matrix. In the rest of the paper, we introduce the Gaussian BP algorithm in Section~\ref{sec2} and discuss the walk summability condition in Section~\ref{sec3}, followed by convergence rate analysis in Section~\ref{sec4}, illustrating examples in Section~\ref{sec5} and conclusions in Section~\ref{sec6}. 

\section{Problem Formulation}\label{sec2}

\begin{figure}[ht]
\begin{center}
\includegraphics[width=5.0cm]{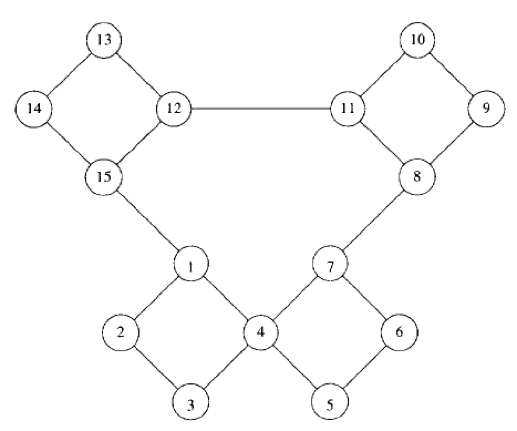}
\end{center}
\caption{An example of Markov Network}\label{fig:0}
\end{figure}

A Gaussian graphical model is a Markov network with Gaussian distributions, characterised by an undirected graph $\mathcal{G}=\{\mathcal{V}, \mathcal{E}\}$, where $\mathcal{V}=\{1, 2, \ldots, n\}$ represents the set of nodes and $\mathcal{E}$ is the set of edges (or unordered pairs $\{i,j\}\subset \mathcal{V}$), with each node $i\in \mathcal{V}$ being associated with a random variable $x_i$. Fig.~\ref{fig:0} shows an example of Markov network. 
The joint probability density for $x=\mathrm{col}\{x_1, x_2, \ldots, x_n\}$ is given by the following Gaussian density function:
\begin{align}
p(x)\propto \exp\{-\frac{1}{2}x^TAx+b^Tx\}, \label{eq:p}
\end{align}
where $A=\{a_{ij}\}$ is a sparse {\em information matrix} with $a_{ij}=0$ for all $\{i,j\}\not\in\mathcal{E}$, which is a symmetric and positive definite matrix, and $b$ is the {\em potential vector}.  It is straightforward to verify that the mean vector $\mu=\mathbb{E}\{x\}$ and covariance matrix $P=\mathbb{E}\{(x-\mu)(x-\mu)^T\}$ are given, respectively, by 
\begin{align}
\mu=A^{-1}b, \ \ P=A^{-1}. \label{eq:p1}
\end{align}

The problem of concern is for each node $i\in \mathcal{V}$ to compute, in a distributed fashion, the marginal density function $p_i(x_i)$ of $x_i$, defined by 
\begin{align}
p_i(x_i) = \int_{x_{-i}} p(x) dx_{-i},\label{eq:p2}
\end{align}
where $x_{-i}$ denotes the vector $x$ with the component $x_i$ removed.  It is well known that this amounts to computing the {\em marginal mean} $\mu_i$ (the $i$-th term of $\mu$) and {\em marginal variance} $p_{ii}$ (the $i$-th diagonal term of $P$). 

Using the Gaussian graphical model, $p(x)$ can be factorised into
\begin{align*}
p(x) \propto \prod_{i\in \mathcal{V}} \phi_i(x_i) \prod_{\{i,j\}\in \mathcal{E}} \phi_(x_i,x_j)
\end{align*}
with 
\begin{align*}
\phi_i(x_i) &= \exp\{-\frac{1}{2}a_{ii}x_i^2+b_ix_i\}, \\
\phi_{ij}(x_i,x_j)&=\exp\{-a_{ij}x_ix_j\}.
\end{align*}
The BP algorithm is an iterative message-passing algorithm for computing $p_i(x_i)$. In each iteration $k$, each node $i\in \mathcal{V}$  computes and transmits to each node $j\in \mathcal{N}_i$ (the set of neighbouring nodes of $i$)  the {\em message} $m_{i\rightarrow j}^{(k)}(x_j)$:
\begin{align*}
m_{i\rightarrow j}^{(k)}(x_j) = \int \phi_{ij}(x_i,x_j)\phi_i(x_i)\prod_{v\in \mathcal{N}_i\backslash j} m_{v\rightarrow i}^{(k-1)}(x_i)dx_i,
\end{align*}
where $m_{v\rightarrow i}^{(k-1)}$ is the message node $i$ receives from its neighbouring node $v$ in iteration $k-1$. 
This results in the marginal density $p_i(x_i)$ to be estimated in iteration $k$ as 
\begin{align*}
p_i^{(k)}(x_i)\propto \phi_i(x_i)\prod_{v\in \mathcal{N}_i} m_{v\rightarrow i}^{(k-1)}(x_i).
\end{align*}
For a Gaussian graphical model, the message $m_{i\rightarrow j}^{(k)}(x_j)$ can be expressed as
\begin{align*}
m_{i\rightarrow j}^{(k)}(x_j) \propto \exp\{-\frac{1}{2} \mathbf{a}_{i\rightarrow j}(k)x_j^2 + \mathbf{b}_{i\rightarrow j}(k)x_j\}.
\end{align*}
 This results in the Gaussian BP algorithm below:
\begin{align*}
\mathbf{a}_{i\rightarrow j}(k)&= -\frac{a_{ij}a_{ji}}{a_{i\rightarrow j}(k)}, \ \ \mathbf{b}_{i\rightarrow j}(k)=-\frac{a_{ji}b_{i\rightarrow j}(k)}{a_{i\rightarrow j}(k)} 
\end{align*}
with
\begin{align*}
a_{i\rightarrow j}(k) &= a_{ii} + \sum_{v\in \mathcal{N}_i\backslash j} \mathbf{a}_{v\rightarrow i}(k-1), \\
b_{i\rightarrow j}(k) &= b_i + \sum_{v\in \mathcal{N}_i\backslash j} \mathbf{b}_{v\rightarrow i}(k-1).
\end{align*}
The initialisation is done by taking $a_{i\rightarrow j}(0)=a_{ii}$ and $b_{i\rightarrow j}(0)=b_i$. 
The marginal mean and marginal variance of $p_i^{(k)}(x_i)$ are then given by, respectively, 
\begin{align}
\mu_i(k) &= \frac{b_i+\sum_{v\in \mathcal{N}_i} \mathbf{b}_{v\rightarrow i}(k-1)}{a_{ii}+ \sum_{v\in \mathcal{N}_i} \mathbf{a}_{v\rightarrow i}(k-1)}, \label{eq:BPmu}\\
 p_{ii}(k)&= \frac{1}{a_{ii}+ \sum_{v\in \mathcal{N}_i} \mathbf{a}_{v\rightarrow i}(k-1)}.\label{eq:BPP}
\end{align}

It is well known \cite{Weiss} that, when the graph $\mathcal{G}$ is acyclic, the Gaussian BP algorithm above converges in $d$ iterations with $\mu_i(k)=\mu_i$ and $p_{ii}(k)=p_{ii}$ for all $i$, where $d$ is the diameter of $\mathcal{G}$ (i.e., the largest distance between any two nodes in $\mathcal{G}$). Actually, for for each node $i$, $d_i$ iterations are sufficient to yield the above convergence, where $d_i$ is the largest distance from any node in $\mathcal{G}$ to node $i$ \cite{Weiss}. (The distance of two nodes is the minimum path length between the nodes.)   

For cyclic (or loopy) graphs, the Gaussian BP algorithm produces the correct marginal means asymptotically under certain conditions. In particular, it has been established in \cite{Mailoutov} that $\mu_i(k)$ converges to $u_i$ for all $i$ asymptotically under the so-called {\em walk summability} condition. This condition is also known to be equivalent to requiring the matrix $A$ to be {\em generalised diagonally dominant} \cite{Boman}. 

The goal of this paper is to study the convergence rate of the Gaussian BP algorithm under the same walk summability  condition. 

\section{Walk Summability}\label{sec3}
Walk-sum analysis is an elegant approach introduced in \cite{Mailoutov} (and their earlier references thereof) for studying the convergence of the Gaussian BP algorithm. Here we provide a quick summary of this approach.

Given a matrix $R=\{r_{ij}\}\in \mathbb{R}^{n\times n}$ and its induced graph $\mathcal{G}=(\mathcal{V},\mathcal{E})$, a {\em walk} $w$ in the graph is a node sequence 
\begin{align}
w=(w_{0},w_{1},\cdots, w_{l}), \ \forall \ w_{i}\in \mathcal{V}, (w_{i}, w_{i+1})\in \mathcal{E}, \label{eq:walk}
\end{align}
and its {\em length} is $l$. The {\em weight} of the walk is defined to be 
\begin{align}
\phi(w)=\prod_{i=0}^{l-1}r_{w_{i}w_{i+1}}. \label{eq:weight}
\end{align}
As a convention, a single node $i\in \mathcal{V}$ is regarded as a special (zero-length) walk with its weight $\phi(i)=1$. A walk $w$ from node $i$ to $j$ is also denoted by $w: i\rightarrow j$, and such a walk with length $l$ is denoted by $w: i^{\underrightarrow{l}}j$.  The set of all walks from node $i$ to node $j$ is denoted by $\{i \rightarrow j\}$, and the set of all length-$l$ walks from node $i$ to node $j$ is denoted by $\{i^{\underrightarrow{l}}j\}$. The {\em walk-sum} of a set of weights $W$ is denoted by $\phi(W) = \sum_{w\in W}\phi(w)$. 

The importance of walk sums is revealed in the relationship that $(i,j)$-th element of matrix $R^{l}$ is equal to:
\begin{align}
(R^{l})_{ij}&=\sum_{w_{1},\cdots,w_{l-1}}r_{iw_{1}}r_{w_{1}w_{2}}\cdots r_{w_{l-1}j}=\sum_{w:i^{\underrightarrow{l}}j}\phi(w), \label{eq:sum}
\end{align}
which can be verified by matrix multiplication. Now we give the definition of walk summability~\cite{Mailoutov}. 

\begin{defn}\label{def:1}
A matrix $A\in \mathbb{R}^{n\times n}$ with $a_{ii}=1$ for all $i$ is said to be {\em walk-summable} if all the walk-sums $\phi(\{i\rightarrow j\})$ {\em converge absolutely}, i.e.,  $\sum_{w:i\rightarrow j} |\phi(w)|$ converges for all $i, j$.  This is the same as the unordered sum $\sum_{w:i\rightarrow j}\phi(w)$ is well defined (i.e., converges to the same value for every possible summation order) for all $i, j$.  Further, a linear system $Ax=b$ is said to be {\em walk-summable} if $A$ is walk-summable. 
\end{defn}

Defining $R=\{r_{ij}\}=I-A$ and $\bar{R}=\{|r_{ij}|\}$, the following properties are known for walk-summable systems~\cite{Mailoutov}.
 
\begin{lem}\label{lem:1}
The following conditions are equivalent. 
\begin{itemize}
\item $A\in \mathbb{R}^{n\times n}$ with $a_{ii}=1$ for all $i$ is walk-summable;
\item $\sum_{l}\bar{R}^{l}$ converges;
\item $\rho(\bar{R})<1$;
\item $I-\bar{R}\succ 0$.
\end{itemize}
In addition, $\rho(R)\le \rho(\bar{R})$. 
\end{lem}

Using the walk-sum interpretation, the Gaussian variance $P$ and mean $\mu$ in (\ref{eq:p1}) can be expressed by walk sums under the assumption of walk summability~\cite{Mailoutov}. More specifically, using
\begin{equation}
P=A^{-1}=(I-R)^{-1}=\sum_{l=0}^{\infty}R^{l} 
\end{equation}
and (\ref{eq:sum}), we get 
\begin{align}
P_{ij}&=\sum_{l=0}^{\infty}(R^{l})_{ij}=\sum_{l=0}^{\infty}\sum_{w:i^{\underrightarrow{l}}j}\phi(w)=\sum_{w:i\rightarrow j}\phi(w). \label{eq:pij}
\end{align}
Similarly, using $\mu=A^{-1}b=\sum_{l=0}^{\infty}R^lb$, we get
\begin{align}
\mu_{i}&=(\sum_{l=0}^{\infty}R^{l}b)_{i}\nonumber \\
&=\sum_{j=1}^{n}\sum_{l=0}^{\infty}(R^{l})_{ij}b_{j}\nonumber \\
&=\sum_{j=1}^{n}\sum_{l=0}^{\infty}\sum_{w:j^{\underrightarrow{l}}i}\phi(w)b_{j} \nonumber \\
&=\sum_{j=1}^{n}\sum_{w:j\rightarrow i}\phi(w)b_{j}. \label{eq:mui}
\end{align}
 
 The connection between walk summability and diagonal dominance is revealed in the result below \cite{Boman}. Recall \cite{Saad} that a matrix $A=\{a_{ij}\}$ is called {\em diagonally dominant} if $a_{ii}>0$ and $a_{ii} >\sum_{j\ne i}|a_{ij}|$ for all $i$. 
 \begin{lem}\label{lem:1-1}
A matrix $A\in\mathbb{R}^{n\times n}$ with $a_{ii}=1$ for all $i$ is walk-summable (i.e., $\rho(\bar{R})<1$) if and only if $A$ is {\em generalised diagonally dominant}, i.e., there exists a diagonal matrix $D>0$ such that $D^{-1}AD$ is diagonally dominant. 
 \end{lem}
 
\section{Convergence Rate Analysis}\label{sec4}

This section presents the main result of this paper on the convergence rate of the Gaussian BP algorithm. The key to this analysis is the so-called {\em unwrapped tree graph} proposed in \cite{Weiss}, which is a computation tree graph, associated with the Gaussian BP algorithm. Using this tool, the asymptotic convergence of the Gaussian BP algorithm was proved in \cite{Weiss} under the assumption of diagonal dominance. This tool was further used in \cite{Mailoutov} to relax the diagonal dominance assumption to walk summability (or equivalently, generalised diagonal dominance).  Here we use the same tool for  convergence rate analysis. 

\subsection{Unwrapped Tree Graph}

Following the work of~\cite{Weiss}, we construct an {\em unwrapped tree} with {\em depth} $t>0$ for a loopy graph $\mathcal{G}$~\cite{Weiss}. Take node $i$ to be the root and then iterate the following procedure $t$ times:
\begin{itemize}
\item Find all leaves of the tree (start with the root);
\item For each leaf, find all the nodes in the loopy graph that neighbor this leaf node, except its parent node in the tree, and add all these node as the children to this leaf node.
\end{itemize}

\begin{figure}[ht]
\begin{picture}(240,135)
\put(54,124){\circle{6}}
\put(57,123){\line(1,-1){39}} 
\put(51,123){\line(-1,-1){39}} 
\put(54,121){\line(0,-1){38}} 
\put(54,80){\circle{6}}
\put(12,81){\circle{6}}
\put(96,81){\circle{6}}
\put(54,77){\line(0,-1){38}}
\put(56,38){\line(1,1){40}} 
\put(52,38){\line(-1,1){40}} 
\put(54,36){\circle{6}}
\put(58,30){$5$}
\put(58,126){$1$}
\put(58,75){$3$}
\put(17,75){$2$}
\put(99,75){$4$}

\put(176,134){\circle{6}}
\put(179,133){\line(1,-1){39}} 
\put(173,133){\line(-1,-1){39}} 
\put(176,131){\line(0,-1){38}} 
\put(176,90){\circle{6}}
\put(134,91){\circle{6}}
\put(218,91){\circle{6}}
\put(176,87){\line(0,-1){25}}
\put(134,88){\line(0,-1){25}}
\put(218,88){\line(0,-1){25}}
\put(134,60){\circle{6}}
\put(176,60){\circle{6}}
\put(218,60){\circle{6}}
\put(180,136){$1$}
\put(180,85){$3$}
\put(139,85){$2$}
\put(221,85){$4$}
\put(139,60){$5$}
\put(180,60){$5'$}
\put(222,60){$5''$}
\put(133,57){\line(-3,-5){12}}
\put(135,57){\line(3,-5){12}}
\put(175,57){\line(-3,-5){12}}
\put(177,57){\line(3,-5){12}}
\put(217,57){\line(-3,-5){12}}
\put(219,57){\line(3,-5){12}}
\put(121,34){\circle{6}}
\put(147,34){\circle{6}}
\put(163,34){\circle{6}}
\put(190,34){\circle{6}}
\put(205,34){\circle{6}}
\put(231,34){\circle{6}}
\put(121,31){\line(0,-1){24}}
\put(147,31){\line(0,-1){24}}
\put(163,31){\line(0,-1){24}}
\put(190,31){\line(0,-1){24}}
\put(205,31){\line(0,-1){24}}
\put(231,31){\line(0,-1){24}}
\put(121,4){\circle{6}}
\put(147,4){\circle{6}}
\put(163,4){\circle{6}}
\put(190,4){\circle{6}}
\put(205,4){\circle{6}}
\put(231,4){\circle{6}}
\put(110,32){$4'$}
\put(137,32){$3'$}
\put(167,32){$2'$}
\put(177,32){$4''$}
\put(209,32){$3''$}
\put(235,32){$2''$}
\put(110,2){$1'$}
\put(136,2){$1''$}
\put(166,2){$1'''$}
\put(179,2){$1`$}
\put(209,2){$1``$}
\put(235,2){$1```$}
\end{picture}
  \caption{Left: A loopy graph. Right: The unwrapped tree for root node 1 with 4 layers ($t=4$)}\label{fig:2}
\end{figure}
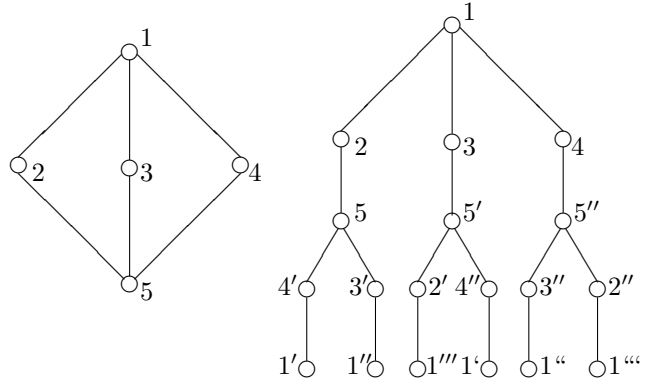

The variables and weights for each node in the unwrapped tree are copied from the corresponding nodes in the loopy graph. 
It is clear that taking each node as root node will generate a different unwrapped tree.  Fig.~\ref{fig:2} shows the unwrapped tree around root node 1 for a loopy graph. Note, for example, that nodes $1', 1'',1''', 1`, 1``,1```$ all carry the same values $b_1$ and $a_{11}$. Similarly, if node 1' is the parent (or child) of node $j'$ in the unwrapped tree, and node 1 and node $j$ are a wrapped version of nodes 1 and $j$, then $a_{1'j'}=a_{1j}$ (or $a_{j'1'}=a_{j1}$).  A similar comment applies to unwrapped $b_i$. 

List the nodes in the unwrapped tree in {\em breadth first} order, by starting from the root node, followed by the first layer (i.e., the children of the root node), then the second layer, etc. Denote the unwrapped tree as $\mathcal{G}_i^{(t)}=\{\mathcal{V}_i^{(t)}, \mathcal{E}_i^{(t)}\}$ with the associated matrix $A_i^{(t)}=I-R_i^{(t)}$ and vector $b_i^{(t)}$.  It is obvious that $\mathcal{G}_i^{(t)}$ is connected by construction.  

We have the following key property.

\begin{lem}\label{lem:loopy1}
  \cite{Mailoutov} There is a one-to-one correspondence between finite-length walks in $\mathcal{G}$ that end at $i$, and walks in $\mathcal{G}_i^{(\infty)}$.  That is, every finite-length walk in $\mathcal{G}$ has a counterpart in some $\mathcal{G}_i^{(k)}$ with some $i\in \mathcal{V}$ and some sufficiently large $k$, and every finite-length walk in $\mathcal{G}_i^{(k)}$ for any $i\in \mathcal{V}$ and $k\ge0$ corresponds to a finite-length walk in $\mathcal{G}$.  
\end{lem}

\subsection{Main Result}
We first establish a relationship between $\mu_i(k)$ in (\ref{eq:BPmu}) (obtained by Gaussian BP) and the walks in $\mathcal{G}_i^{(k)}$.

\begin{lem}\label{lem:2}
Under the assumption that the information matrix $A$ in (\ref{eq:p}) is walk summable, we have, for any $i\in \mathcal{V}$ and $k\ge0$, 
\begin{align}
\mu_i(k) = \sum_{j=1}^n \sum_{w: j\rightarrow i  |\mathcal{G}_i^{(k)}} \phi(w)b_j \label{eq:mui2}
\end{align}
where $w: j\rightarrow i  |\mathcal{G}_i^{(k)}$ denotes a walk  from $j$ to $i$ inside the unwrapped graph $\mathcal{G}_i^{(k)}$. 
\end{lem}

\begin{IEEEproof}
Without loss of generality, we assume $i=1$.  For the unwrapped graph $\mathcal{G}_1^{(k)}$, consider the corresponding matrix $A_1^{(k)}$ and vector $b_1^{(k)}$. Define $z^{(k)}=(A_1^{(k)})^{-1}b_1^{(k)}$. Then, $z^{(k)}$ can be solved by applying the Gaussian BP algorithm on $\mathcal{G}_1^{(k)}$.  As noted in Section~\ref{sec2}, since $\mathcal{G}_1^{(k)}$ is a tree graph, it is well known \cite{Weiss} that applying Gaussian BP to $\mathcal{G}_1^{k}$ results in a correct solution for $z_1^{(k)}$ (the first component of $z^{(k)}$) in $k$ iterations because every node in $\mathcal{G}_1^{(k)}$ is no more than $k$ hops away from node 1. On the other hand, due to the fact that the parameters in $A_1^{(k)}$ and $b_1^{(k)}$ are all copied from $A$ and $b$, applying Gaussian BP to the original graph $\mathcal{G}$ for $k$ iterations is identical to applying it to $\mathcal{G}_1(k)$. That is, $\mu_1^{(k)}$ in (\ref{eq:BPmu}), which is obtained by applying Gaussian BP on $\mathcal{G}$ for $k$ iterations, is equal to $z_1^{(k)}$.  Now, it is also known that every tree graph is walk-summable \cite{Mailoutov}.  Thus, we can apply (\ref{eq:mui}) to $\mathcal{G}_1^{(k)}$ to obtain 
\begin{align*}
z_1^{(k)} &= \sum_{j} \sum_{j\rightarrow 1|\mathcal{G}_1^{(k)}} \phi(w)b_j. 
\end{align*}
Using $z_1^{(k)}=\mu_1(k)$, we have proved (\ref{eq:mui2}) for $i=1$. Hence, the result in the lemma holds. 
\end{IEEEproof}

Now we can state the main result. 

\begin{thm}\label{thm:1}
Suppose the information matrix $A$ in (\ref{eq:p}) is walk-summable. Then, the convergence rate of Gaussian BP algorithm is at least $\rho(\bar{R})$, i.e., 
\begin{align}
|\mu_i(k)-\mu_i|\le \rho(\bar{R})C, \label{eq:rate}
\end{align}
for all $i\in \mathcal{V}$ and $k\ge0$, where $C$ is a constant (independent of $k$). 
\end{thm}
\begin{IEEEproof}
Note from (\ref{eq:p1}) that $\mu=A^{-1}b$. Using (\ref{eq:mui}) and walk summability assumption, we get 
\begin{align*}
\mu_i &=\sum_{j=1}^{n}\sum_{l=0}^{\infty}\sum_{w:j^{\underrightarrow{l}}i}\phi(w)b_{j}.
\end{align*}
Combining it with (\ref{eq:mui2}), we get
\begin{align*}
\mu_i(k)-\mu_i &= \sum_{j=1}^n \sum_{l=0}^{\infty} \left (\sum_{w:j^{\underrightarrow{l}}i|\mathcal{G}_i^{(k)}} \phi(w)b_j - \sum_{w:j^{\underrightarrow{l}}i} \phi(w)b_j \right ).
\end{align*}
Denote by $W_{i}(k)$ the set of all the walks that end at node $i$ with walk length greater than $k$, and by $\tilde{W}_i(k)\subset W_i(k)$ the subset of all the walks containing nodes {\em not} in $\mathcal{G}_k$.  It is clear that every walk in $\tilde{W}_i(k)$ has length greater than $k$. Then, the above expression can be rewritten as 
\begin{align*}
\mu_i(k) - \mu_i &=\sum_{j=1}^n \sum_{l=0}^{\infty} \sum_{w:j^{\underrightarrow{l}}i|\tilde{W}_i^{(k)}} \phi(w)b_j\\
&=\sum_{j=1}^n \sum_{l=k+1}^{\infty} \sum_{w:j^{\underrightarrow{l}}i|\tilde{W}_i^{(k)}} \phi(w)b_j.
 \end{align*}
It follows that
\begin{align*}
|\mu_i(k) - \mu_i| &\le\sum_{j=1}^n \sum_{l=k+1}^{\infty} \sum_{w:j^{\underrightarrow{l}}i|\tilde{W}_i^{(k)}} |\phi(w)||b_j|\\
&\le \sum_{j=1}^n \sum_{l=k+1}^{\infty} \sum_{w:j^{\underrightarrow{l}}i} |\phi(w)||b_j| \\
&= \sum_{j=1}^n \sum_{l=k+1}^{\infty} (\bar{R}^{l})_{ij}|b_j| \\
&= (\sum_{l=k+1}^{\infty} \bar{R}^l|b|)_i \\
&=(\bar{R}^{k} \sum_{l=1}^{\infty} \bar{R}^l|b|)_i \\
&\le \rho(\bar{R})^k C,
 \end{align*}
where $C=\max_i (\sum_{l=1}^{\infty} \bar{R}^l|b|)_i$ is bounded due to the fact that $\rho(\bar{R})<1$. Hence, (\ref{eq:rate}) holds for all $i$.
\end{IEEEproof}

\section{Illustrating Examples}\label{sec5}
\begin{figure}[ht]
\begin{center}
\includegraphics[width=5.0cm]{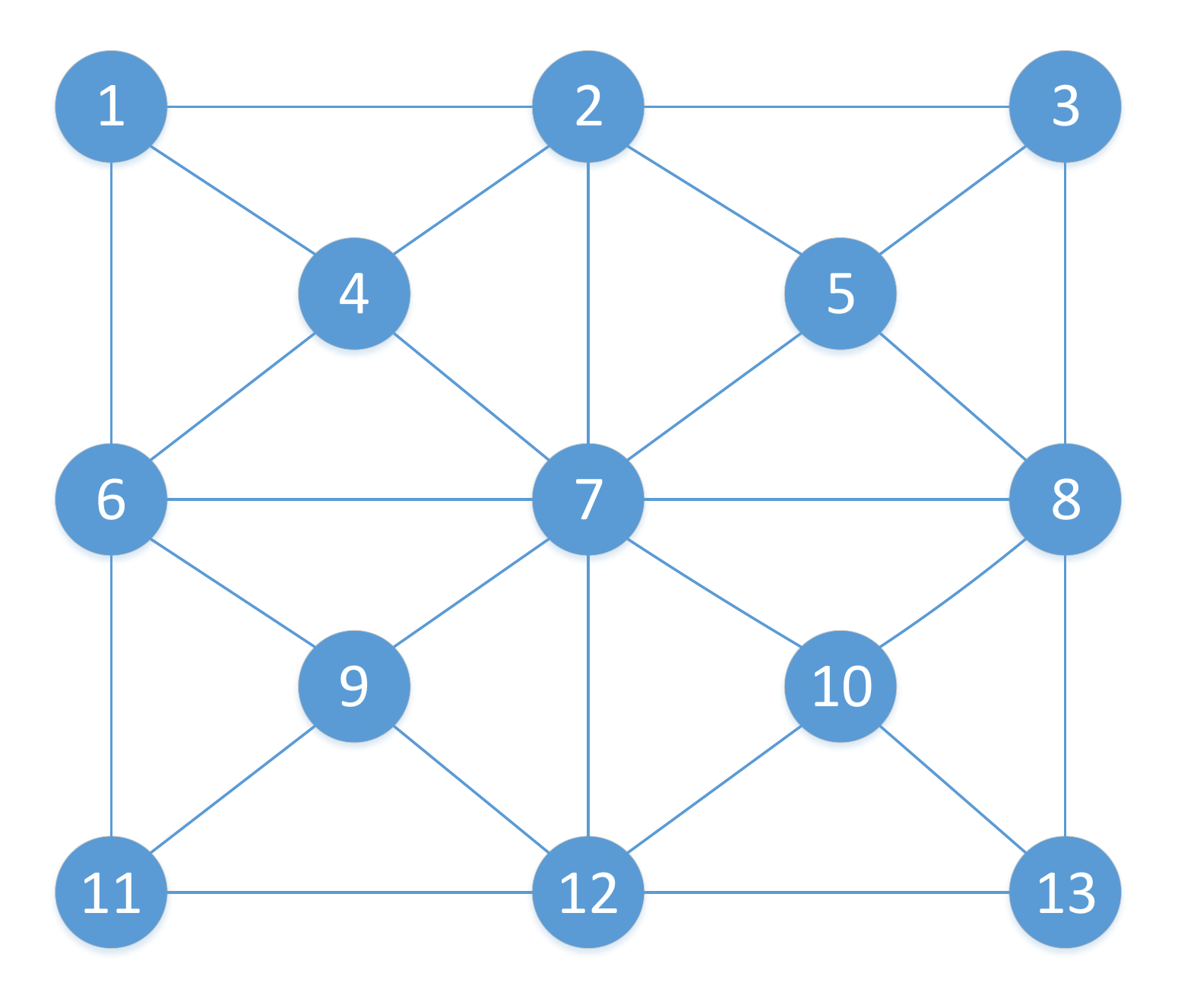}
\end{center}
\caption{13-node graph}\label{fig:s1}
\end{figure}

\begin{figure}[ht]
\begin{center}
\includegraphics[width=8.5cm]{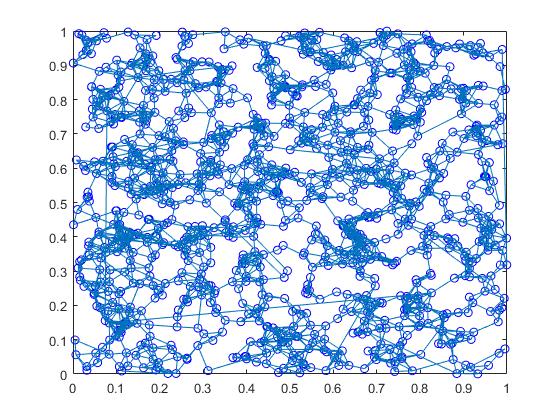}
\end{center}
\caption{1000-node graph}\label{fig:s2}
\end{figure}

To illustrate the convergence rate bound in Theorem~\ref{thm:1}, we give two loopy graphs in this section as examples. The first one is a 13-node graph with at most 5 neighbouring nodes for each node, as shown in Fig.~\ref{fig:s1}. The second example is a 1000-node graph with at most $6$ randomly selected neighbouring nodes for each node, as shown in Fig.~\ref{fig:s2}. In both cases, the resulting matrix $A$ is sparse. The parameters of the corresponding Gaussian density function $p(x)\propto \exp \{-\frac{1}{2}x^{T}Ax+b^{T}x\}$ are chosen as follows: In the first example, each $a_{ij}$ is a random value in $(-0.26,0.26)$, and in the second example, $a_{ij}$ belongs to $(-0.165,0.165)$ randomly. Additionally, $a_{ii}=1$ and $b_{i}=i,$ $i=1,\cdots, n$. The values of $a_{ij}$ are chosen to ensure diagonal dominance, which in turn ensure walk summability. 

\begin{figure}[ht]
\begin{center}
\includegraphics[width=8.5cm]{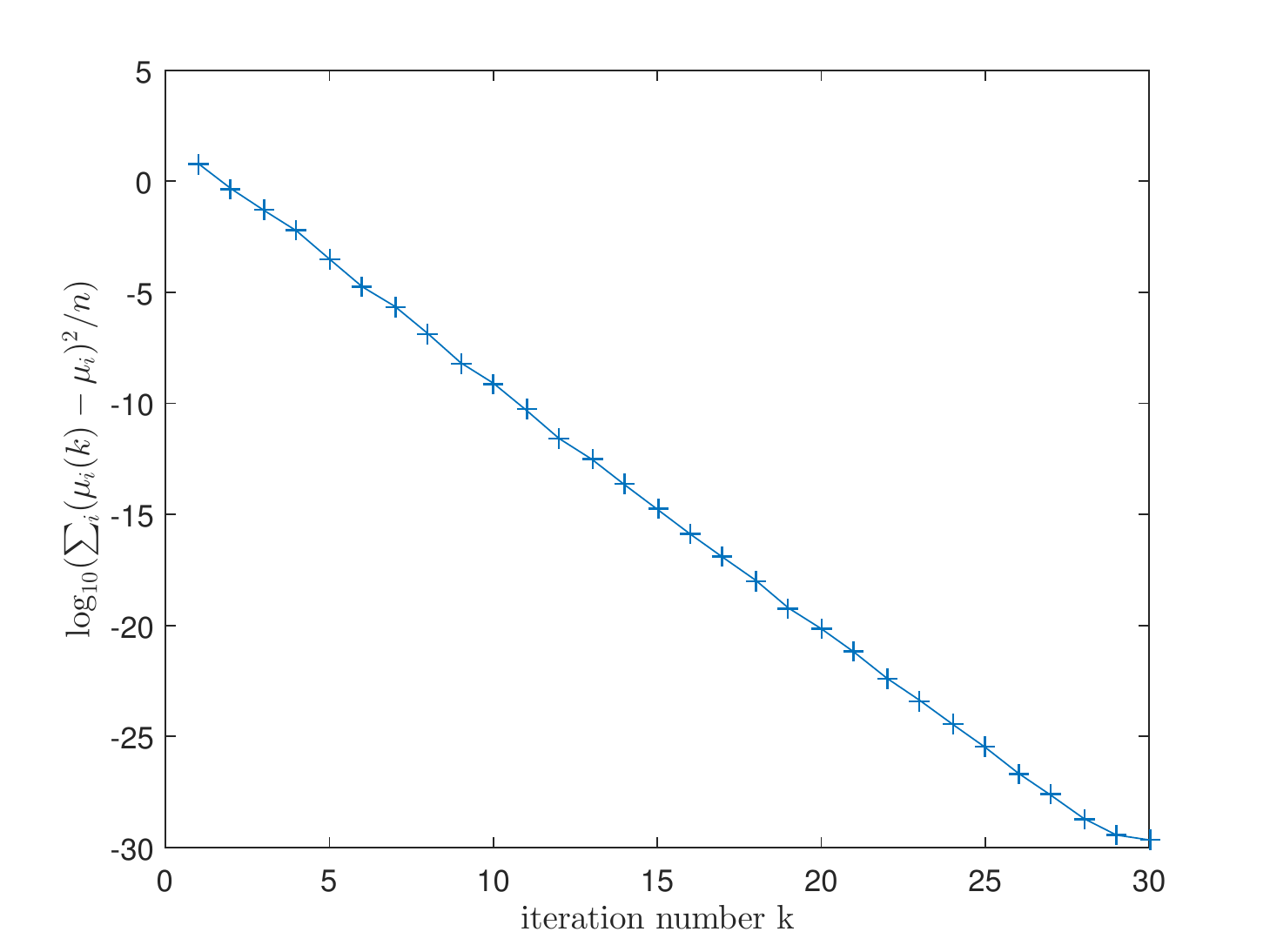}
\end{center}
\caption{Gaussian BP iterations for the 13-node graph}\label{fig:s3}
\end{figure}

\begin{figure}[ht]
\begin{center}
\includegraphics[width=8.5cm]{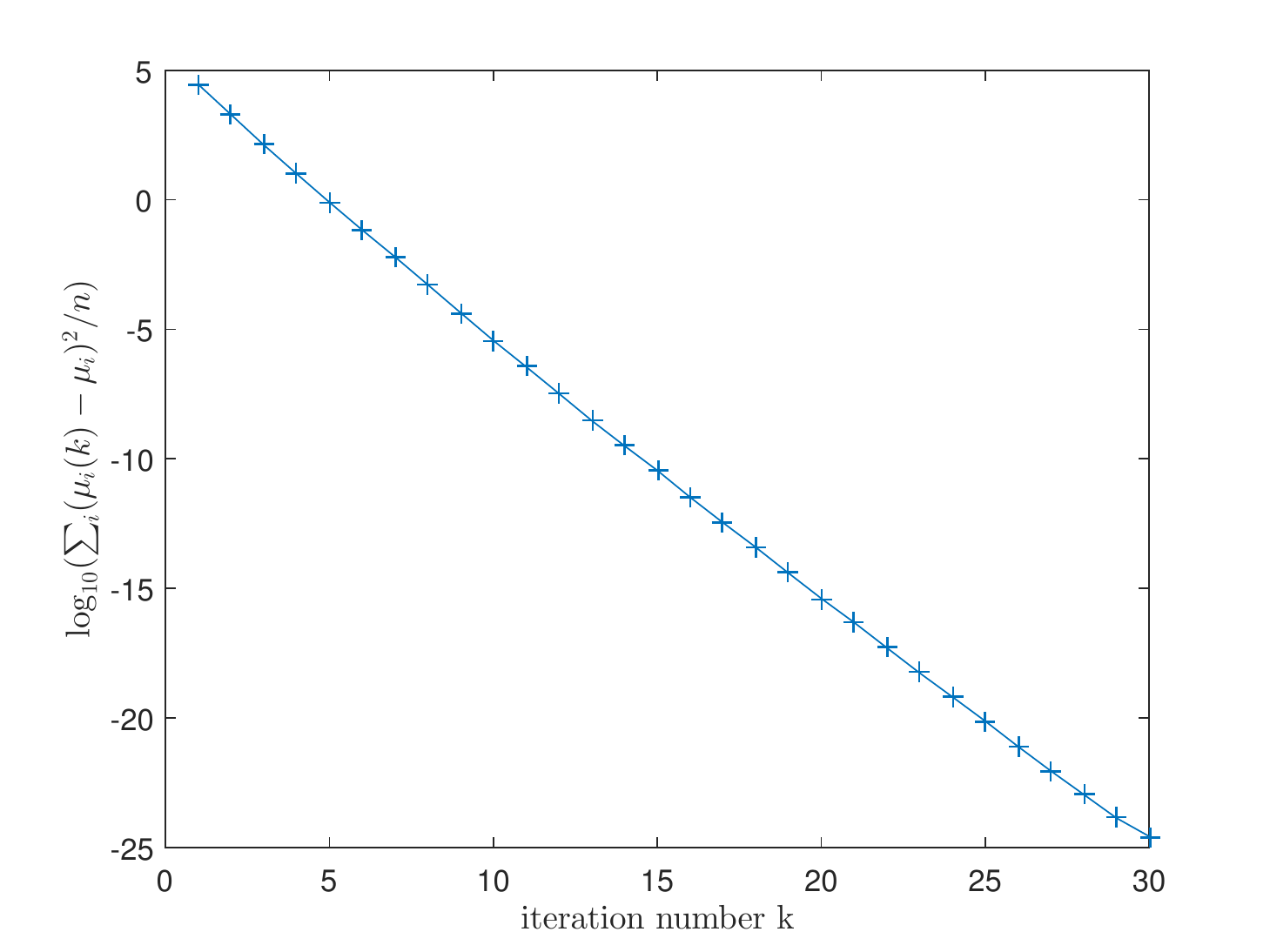}
\end{center}
\caption{Gaussian BP iterations for the 1000-node graph}\label{fig:s4}
\end{figure}

The Gaussian BP simulation results are shown in Figs.~\ref{fig:s3}-\ref{fig:s4}. The $x$-axis stands for iteration numbers and the $y$-axis stands for a $\log$ form of error between the true Gaussian mean $\mu$ and its estimate $\mu(k)$ calculated by the Gaussian BP algorithm, or more precisely, $\log_{10}(\sum_{i}(\mu_i(k)-\mu_i)^2/n)$. The simulation results for both two examples have shown that the error decreases exponentially with the increase of the iteration number. The slope for the 13-node example is measured to be roughly -1.0502, corresponding to the convergence rate of $10^{-1.0502/2} \approx 0.2985$. The slope for the 1000-node example is measured to be roughly -1.0642, corresponding to the convergence rate of $10^{-1.0642/2} \approx 0.2937$. In comparison, for the 13-node graph, the spectral radius of $\bar{R}$ is $0.6100$; and for the 1000-node graph, the spectral radius of $\bar{R}$ is $0.9671$. We see that in both examples, the $\rho(\bar{R})$ upper bounds the actual convergence rate of the Gaussian BP algorithm. 


\section{Conclusions}\label{sec6}
In this paper, we have analysed the convergence property of the Gaussian BP algorithm for Markov networks and provided a simple bound on the convergence rate. The bound is characterised by $\rho(\bar{R})$ and is guaranteed to be less than 1 under the walk summability (or generalised diagonal dominance) assumption. This result gives a simple extension to the known asymptotic convergence property of the Gaussian BP algorithm under the same assumption \cite{Weiss,Mailoutov}. We see in the simulation results that the actual convergence rate is faster than predicted by $\rho(\bar{R})$. It would be interesting to see how this bound can be further improved. Other future directions include relaxing the walk summability assumption, and generalising the Gaussian BP algorithm to wider distributed estimation and distributed optimisation problems.

\end{document}